\documentclass[11pt]{camel-ai}

\usepackage[toc,page,header]{appendix}

\usepackage[utf8]{inputenc} 
\usepackage[T1]{fontenc}    
\usepackage{hyperref}       
\usepackage{url}            
\usepackage{booktabs}       
\usepackage{amsfonts}       
\usepackage{nicefrac}       
\usepackage{microtype}      
\usepackage{xspace}
\usepackage{bm}
\usepackage{bbm}
\usepackage{tabularx}
\usepackage{amssymb}
\usepackage{amsmath}
\usepackage{mathtools}
\usepackage{amsthm}
\usepackage{pifont} 
\usepackage{multirow}
\usepackage{makecell}
\usepackage{paralist}
\usepackage{color}
\usepackage{colortbl}
\usepackage{adjustbox}
\usepackage[edges]{forest}
\usepackage{amssymb}
\usepackage{pifont}
\usepackage{wrapfig}
\usepackage[noend]{algpseudocode}
\usepackage{algorithm}
\usepackage{algorithmicx}
\usepackage{algpseudocode}
\usepackage{pifont}
\usepackage{pgfplots}
\pgfplotsset{compat=1.17}
\usepackage{pgfplotstable}
\usepackage{svg}

\usepackage{tikz}

\usepgfplotslibrary{groupplots}

\usepackage{caption}
\usepackage{graphicx}
\usepackage[most]{tcolorbox}
\tcbuselibrary{breakable}
\newtcolorbox{mycustombox}[1]{
  enhanced,
  colback=black!5!white,
  colframe=black!75!white,
  boxrule=0.4pt,
  coltitle=white,
  title=#1,
  titlerule=0.4pt,
  fontupper=\small,
  fonttitle=\small,
  before upper={\par\smallskipamount},
  breakable,
  left=3mm,
  right=3mm,
  top=2mm,
  bottom=2mm,
  boxsep=1mm,
}

\definecolor{mygray}{gray}{0.9}
\definecolor{newgreen}{RGB}{78, 173, 102}
\definecolor{speciallink}{HTML}{5a2afd}

\definecolor{mycolor_green}{HTML}{E6F8E0}
\definecolor{mmada_color}{HTML}{EFF7FF}
\definecolor{mycolor_blue}{HTML}{E7EFFA}
\definecolor{mycolor_green}{HTML}{E6F8E0}
\definecolor{mycolor_gray}{HTML}{ECECEC}
\definecolor{pearDark}{HTML}{2980B9}
\definecolor{demphcolor}{RGB}{90,42,253}
\definecolor{mydarkgreen}{RGB}{0,100,0}

\usepackage{textcomp}  
\usepackage{scalerel}  

\usepackage{times}

\usepackage[utf8]{inputenc}
\usepackage{microtype}
\usepackage{graphicx}
\usepackage{booktabs}
\usepackage{hyperref}
\usepackage{stackengine}
\usepackage{wrapfig}
\usepackage{etoolbox}
\usepackage{fancyvrb}
\usepackage{multirow}
\usepackage{amssymb}
\usepackage{adjustbox}
\usepackage{algorithm}
\usepackage{algorithmicx}
\usepackage{algpseudocode}
\usepackage{amsfonts}       
\usepackage{booktabs}       
\usepackage{float}
\usepackage{fnpos}
\usepackage{graphicx}
\usepackage{microtype}      
\usepackage{nicefrac}       
\usepackage[textsize=footnotesize]{todonotes}
\usepackage{xspace}
\usepackage{mathtools}
\usepackage{enumitem}
\usepackage{thmtools}
\usepackage{wrapfig}
\usepackage{bbm}
\usepackage{wrapfig}
\usepackage{transparent}
\usepackage{enumitem}
\usepackage{changepage} 
\usepackage{pythonhighlight}
\usepackage{xurl}
\usepackage{subcaption}
\usepackage{svg}
\usepackage{booktabs}

\usepackage[most]{tcolorbox}
\usetikzlibrary{shadows}
\definecolor{mine}{RGB}{205, 232, 248}%
\definecolor{minedark}{RGB}{160, 190, 210}%
\definecolor{revision}{RGB}{210, 22, 123}

\usepackage[]{nomencl}   
    \makenomenclature

\makeatletter
\newcommand\myscriptsize{\@setfontsize\myscriptsize{8pt}{9pt}}
\makeatother

\providetoggle{nomsort}
\settoggle{nomsort}{true} 

\makeatletter
\iftoggle{nomsort}{%
    \let\old@@@nomenclature=\@@@nomenclature        
        \newcounter{@nomcount} \setcounter{@nomcount}{0}%
        \renewcommand\the@nomcount{\two@digits{\value{@nomcount}}}
        \def\@@@nomenclature[#1]#2#3{
          \addtocounter{@nomcount}{1}%
        \def\@tempa{#2}\def\@tempb{#3}%
          \protected@write\@nomenclaturefile{}%
          {\string\nomenclatureentry{\the@nomcount\nom@verb\@tempa @[{\nom@verb\@tempa}]%
          \begingroup\nom@verb\@tempb\protect\nomeqref{\theequation}%
          |nompageref}{\thepage}}%
          \endgroup
          \@esphack}%
      }{}
\makeatother
\usepackage{silence}
\usepackage{tikz}

\tikzstyle{every picture}+=[remember picture]

\usepackage[most]{tcolorbox}
\usetikzlibrary{shadows}
\tcbuselibrary{theorems}

\definecolor{functionclass}{RGB}{113, 153, 194}
\definecolor{lossfunction}{RGB}{119, 221, 119}
\definecolor{regterm}{RGB}{255, 179, 71}

\algnewcommand{\LineComment}[1]{\Statex \(\quad \ \ \textcolor{LightSteelBlue3}{\triangledown} \quad \) \textcolor{LightSteelBlue3}{$/^{\star}$ #1 $^{\star}/$}}

\algnewcommand{\NoIndLineComment}[1]{\Statex \(\textcolor{LightSteelBlue3}{\triangledown} \quad \) \textcolor{LightSteelBlue3}{$/^{\star}$ #1 $^{\star}/$}}

\newcounter{exa}
\definecolor{gblue}{RGB}{66,133,244}
\definecolor{gred}{RGB}{219,68,55}
\definecolor{gyellow}{RGB}{244,180,0}
\definecolor{ggreen}{RGB}{15,157,88}
\definecolor{lpcolor}{RGB}{42,74,138}
\definecolor{morelcolor}{RGB}{185,18,32}
\definecolor{bgcolor}{RGB}{230,245,208}
\definecolor{framecolor}{RGB}{244,109,67}
\definecolor{mulberry}{rgb}{0.77, 0.29, 0.55}
\hypersetup{colorlinks=true,
            citecolor=gblue,
            urlcolor=ggreen,
            linkcolor=purple}
\definecolor{Chocolate3}{RGB}{205, 105, 29}
\definecolor{LightSteelBlue3}{RGB}{162, 181, 205}
\definecolor{DodgerBlue4}{RGB}{16, 78, 139}

\tcbset{
myexample/.style={
  enhanced,
  colback=yellow!10!white,
  colframe=red!50!black,
  fonttitle=\scshape,
  titlerule=0pt,
  title={\refstepcounter{exa}example~\theexa.},
  title style={fill=yellow!10!white},
  coltitle=red!50!black,
  drop shadow,
  highlight math style={reset,colback=LightBlue!50!white,colframe=Navy}
  }
  }
\newtcolorbox{texample}{myexample}

\newtheorem{exampp}{Example}
\usepackage{framed}
\colorlet{shadecolor}{gray!20}

\colorlet{LightLavender}{green!5}
\tcbset{on line, 
        boxsep=4pt, left=0pt,right=0pt,top=0pt,bottom=0pt,
        colframe=white,colback=LightLavender,  
        highlight math style={enhanced}
        }

\allowdisplaybreaks

\usepackage{mathtools}
\usepackage{natbib}
\usepackage{latexsym}

\usepackage{url}
\usepackage{amssymb}
\usepackage[utf8]{inputenc}
\usepackage{microtype}
\usepackage{booktabs}
\usepackage{pifont} 
\usepackage{multirow}
\usepackage{makecell}
\usepackage{paralist}
\usepackage{xspace}
\usepackage{color}
\usepackage{colortbl}
\usepackage{adjustbox}
\usepackage{hyperref} 
\usepackage[edges]{forest}
\usepackage{tikz} 
\usepackage{caption}
\usepackage{amsfonts}

\hypersetup{
    colorlinks,
    linkcolor={blue!80!black},
    citecolor={blue!80!black},
}
\tikzset{
    root/.style =             {align=center, text width=1cm, rounded corners=3pt, line width=0.3mm, fill=gray!10, draw=gray!80, font=\small},
    demographic/.style =         {align=center, text width=1.8cm, rounded corners=3pt, line width=0.3mm, fill=blue!10, draw=blue!80, font=\footnotesize},
    demographic_work/.style =    {align=center, text width=10cm, rounded corners=3pt, line width=0.3mm, fill=blue!10, draw=blue!0, font=\footnotesize},
    character/.style =         {align=center, text width=1.8cm, rounded corners=3pt, line width=0.3mm, fill=red!10, draw=red!80, font=\footnotesize},
    character_work/.style =    {align=center, text width=10cm, rounded corners=3pt, line width=0.3mm, fill=red!10, draw=red!0, font=\footnotesize},
    personalization/.style =           {align=center, text width=1.8cm, rounded corners=3pt, line width=0.3mm, fill=cyan!10, draw=cyan!80, font=\footnotesize},
    personalization_work/.style =      {align=center, text width=10cm, rounded corners=3pt, line width=0.3mm, fill=cyan!10, draw=cyan!0, font=\footnotesize},
    risk/.style =         {align=center, text width=1.8cm, rounded corners=3pt, line width=0.3mm, fill=orange!10, draw=orange!80, font=\footnotesize},
    risk_work/.style =    {align=center, text width=10cm, rounded corners=3pt, line width=0.3mm, fill=orange!10, draw=orange!0, font=\footnotesize},
}

\usepackage{CJK}
\definecolor{myblue}{RGB}{34, 66, 122} 
\renewcommand{\eqref}[1]{Eq.~\ref{#1}}

\usepackage[toc,page,header]{appendix}
\usepackage{minitoc}

\newcommand{\loong}{%
  \includegraphics[height=0.8em]{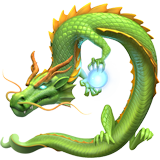}%
  \xspace
}
\newcommand{\loongbench}{\textsc{LoongBench}}

\newcommand{\loongenv}{\textsc{LoongEnv}}

\def\logo{\includegraphics[height=17pt]{assets/logo_loong.png}}
\title{\logo\textls[20]{Loong: Synthesize Long Chain-of-Thoughts at Scale through Verifiers}}

\author[*]{Xingyue Huang}{}
\author[*]{Rishabh}{}
\author[*]{Gregor Franke}{}
\author[*]{Ziyi Yang}{}

\author[$\ddagger$]{Jiamu Bai}{n}
\author[]{Weijie Bai}{}
\author[]{Jinhe Bi}{}
\author[]{Zifeng Ding}{}
\author[]{Yiqun Duan}{}
\author[]{Chengyu Fan}{}
\author[]{Wendong Fan}{}
\author[]{Xin Gao}{}
\author[]{Ruohao Guo}{}
\author[]{Yuan He}{}
\author[]{Zhuangzhuang He}{}
\author[]{Xianglong Hu}{}
\author[]{Neil Johnson}{}
\author[]{Bowen Li}{}
\author[]{Fangru Lin}{}
\author[]{Siyu Lin}{}
\author[]{Tong Liu}{}
\author[]{Yunpu Ma}{}
\author[]{Hao Shen}{}
\author[]{Hao Sun}{}
\author[]{Beibei Wang}{}
\author[]{Fangyijie Wang}{}
\author[]{Hao Wang}{}
\author[]{Haoran Wang}{}
\author[]{Yang Wang}{}
\author[]{Yifeng Wang}{}
\author[]{Zhaowei Wang}{}
\author[]{Ziyang Wang}{}
\author[]{Yifan Wu}{}
\author[]{Zikai Xiao}{}
\author[]{Chengxing Xie}{}
\author[]{Fan Yang}{}
\author[]{Junxiao Yang}{}
\author[]{Qianshuo Ye}{}
\author[]{Ziyu Ye}{}
\author[]{Guangtao Zeng}{}
\author[]{Yuwen Ebony Zhang}{}
\author[]{Zeyu Zhang}{}
\author[]{Zihao Zhu}{}

\author[]{Bernard Ghanem}{n}
\author[]{Philip Torr}{}
\author[$\dagger$]{Guohao Li}{}

\affiliation[] {CAMEL-AI.org}

\contribution[*]{Equal Contribution}
\contribution[\dagger]{Corresponding author}
\contribution[\ddagger]{Authors listed here are in alphabetical order}

\vspace{-0.3cm}
\abstract{
 Recent advances in Large Language Models (LLMs) have shown that their reasoning capabilities can be significantly improved through Reinforcement Learning with Verifiable Reward (RLVR), particularly in domains like mathematics and programming, where ground-truth correctness can be automatically evaluated. However, extending this success to other reasoning-intensive domains remains challenging due to the scarcity of high-quality, verifiable datasets and the high cost of human supervision.
In this work, we introduce the \loong{} Loong Project: an open-source framework for scalable synthetic data generation and verification across a diverse range of reasoning-intensive domains. The framework consists of two key components: (1) \loongbench, a curated seed dataset containing 8,729 human-vetted examples across 12 domains (e.g., Advanced Mathematics, Chemistry, Logic), each paired with executable code and rich metadata; and (2) \loongenv, a modular synthetic data generation environment that supports multiple prompting strategies to produce new question-answer-code triples.
Together, these components form an agent-environment loop that enables reinforcement learning, where an LLM-based agent is rewarded for generating Chain-of-Thought (CoT) solutions that align with code-executed answers.
Empirically, we benchmark \loongbench~on a broad suite of both open-source and proprietary LLMs to evaluate domain coverage and reveal performance bottlenecks. In addition, we conduct a comprehensive analysis of synthetic data generated by \loongenv, examining correctness, difficulty, and diversity. 
Code and documentation are available at \url{https://github.com/camel-ai/loong}.

}

\date{\today}
\correspondence{Guohao Li at \href{mailto:guohao.li@eigent.ai}{\textcolor{speciallink}{guohao.li@eigent.ai}}}
\checkdata[Project Page]{\href{https://github.com/camel-ai/loong}{\textcolor{speciallink}{https://github.com/camel-ai/loong}}}

\begin{document}
\maketitle

\section{Introduction}

Recent Large Reasoning Models such as DeepSeek-R1 \citep{guo2025deepseekr1} and o3~\citep{openai2025o3mini} have demonstrated that the general reasoning capabilities of LLMs greatly improve when base models undergo post-training with Reinforcement Learning (RL) with a verifiable reward \citep{shen2025satori, peng2025graphprm, stojanovski2025reasoninggym}. Mathematics and programming \citep{shen2025satori, wei2022chainofthought} have particularly benefited from this approach, as these domains can be verified quite easily, allowing accurate interpretation of LLM responses and effective comparison to the ground truth on a semantic level. This idea that the ease of verification is crucial to improving domain-specific capabilities has become widely accepted in the research community \citep{raschka2025rlreasoning, rlvrreview2025, ma2025generalreasoner}.

Another critical prerequisite which is often overlooked is the abundance of high-quality datasets, featuring questions paired with verified correct answers in the domains of Maths and Coding \citep{hendrycksmath2021, wei2022chainofthought}. These curated datasets provided the necessary signal for models to learn to construct coherent Chains-of-Thought (CoTs) \citep{wei2022chainofthought}, leading reliably to correct answers.

However, many other domains also require reliable reasoning, such as logic, graph theory, physics, and finance. These domains lack comparable datasets \citep{liu2025synlogic, peng2025graphprm, wu2025synthrl, xie2025finchain}, and human-supervised data production at scale is prohibitively expensive \citep{stojanovski2025reasoninggym, liu2025finr1, xie2025finchain}. Without abundant correct answers to learn from, models cannot easily acquire domain-specific reasoning patterns. This raises a crucial question: \emph{Can similar reasoning performance be achieved in domains beyond maths and programming?}

\paragraph{Approach.}
We present the \loong{} Loong Project: an open framework for scaling synthetic data generation with verifiable supervision across a diverse set of reasoning-centric domains. The framework comprises two key components: 
\begin{enumerate}
\item \loongbench, a meticulously curated seed dataset comprising 8,729 examples across 12 reasoning-intensive domains, each accompanied by executable code and verified answers.
\item \loongenv, a modular and versatile synthetic data generation environment, capable of generating diverse and semantically verifiable question-answer pairs using various automated generation strategies.
\end{enumerate}

\begin{figure}
    \centering
    \includegraphics[width=\linewidth]{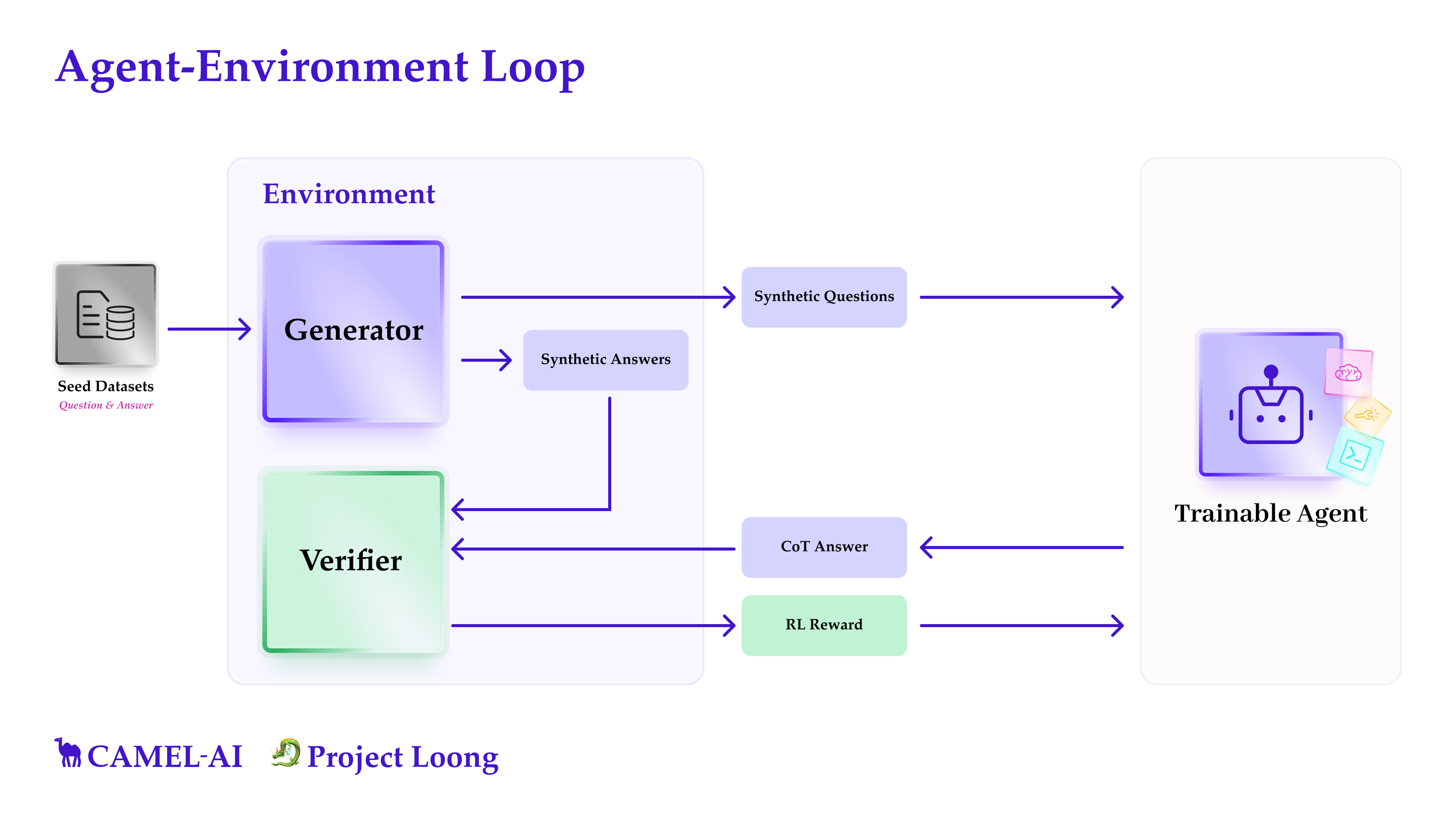}
    \caption{Agent-Environment Loop}
    \label{fig:agent_environment_loop}
\end{figure}

As illustrated in Figure~\ref{fig:agent_environment_loop}, the overall agent-environment loop operates as follows:
First, given a collection of seed datasets, our generator produces synthetic data points consisting of automatically generated questions and corresponding executable codes that answer these questions. Second, these codes are executed within the environment to yield synthetic answers. Third, a trainable agent is prompted to solve the synthetic questions by generating natural language CoT responses. Finally, a verifier compares the agent's CoT-derived answer to the code-generated answer.
This setup will enable large-scale reinforcement learning with minimal human supervision while preserving semantic correctness through automated verification in the future.

\paragraph{Contributions.} Our main contributions are:
\begin{itemize}
    \item We introduce \loongbench, a high-quality seed dataset of 8,729 examples spanning 12 reasoning-intensive domains, each paired with executable code and semantically verified answers.
    
    \item We develop \loongenv, a synthetic data generation environment that supports multiple generation strategies to produce diverse and verifiable question-answer pairs.
    
    \item We benchmark \loongbench~across a diverse suite of large language models-including both open-source and proprietary, general-purpose and reasoning-specialized models-to establish baseline performance and identify domain-specific challenges.

    \item We conduct a detailed analysis of the synthetic data generated by \loongenv, evaluating it in terms of semantic correctness, question difficulty, and diversity.
\end{itemize}

\section{\texorpdfstring{\loong}{\xspace} Loong Project}

The \texorpdfstring{\loong}{\xspace} Loong Project is focused on scaling up synthetic data generation with verification mechanisms across a broad spectrum of domains. We believe that generating synthetic data is essential not just to overcome the lack of datasets in under-represented fields, but also to strengthen reasoning abilities in areas like mathematics and programming by making more training examples available.

Our system relies on a multi-agent setup that starts with a seed dataset and expands it by generating new questions and corresponding answers. These synthetic questions are then passed to a model under training, which attempts to answer them. Verifiers are then employed to compare the model's answers with the pre-generated and pre-computed solutions, checking for semantic agreement.

At the heart of our approach is a simple idea: \emph{an LLM equipped with a code interpreter is often far more reliable when solving complex tasks than one that relies solely on natural language reasoning}. This idea is backed by how many scientific disciplines operate in practice: whether it's physics, neurophysiology, economics, or computational biology, code-based solutions are a standard way to formulate and solve domain-specific problems.

To achieve our goals, two key components are required: a high-quality seed dataset that spans multiple domains, and a modular environment capable of generating synthetic questions and answers in a structured, verifiable manner.

\loongbench{} provides the foundational data for our framework. It consists of 8,729 carefully curated examples covering 12 diverse, reasoning-intensive domains. Each example is annotated with executable code and paired with semantically verified answers. These seed examples ensure coverage of domain-specific patterns while maintaining correctness and diversity, offering a reliable basis for downstream synthetic data generation and benchmarking.

Complementing \loongbench{} is \loongenv{}, a flexible and extensible synthetic data generation environment. \loongenv{} takes the seed examples from \loongbench{} and uses various strategies, including few-shot prompting, self-instruction, and programmatic transformations, to generate new question-answer pairs. It also executes associated code to produce verifiable answers, which are used to supervise and evaluate model performance. This environment is domain-agnostic and modular, supporting plug-and-play verifiers and generation policies across reasoning domains.

\subsection{\loongbench: Human-Vetted Seed Datasets Across Multiple Domains}

\label{sec:seeddataset}

\begin{table}[!t]
\footnotesize
\centering
\caption{Statistics of the \loongbench~by domains.}
\label{tab:seed_dataset}
\begin{tabular}{l|c|c}
\toprule
\textbf{Domain} & \textbf{Main Dependency} & \textbf{Size} \\ \midrule
{Advanced Maths} & \texttt{sympy} & 1,611 \\
{Advanced Physics} & \texttt{sympy, numpy} & 429 \\
{Chemistry} & \texttt{rdkit, numpy} & 3,076 \\
{Computational Biology} & - & 51 \\
{Finance} & \texttt{QuantLib} & 235 \\
{Board Game} & - & 926 \\
{Graph \& Discrete Maths} & \texttt{networkx} & 178 \\
{Logic} & \texttt{python-constraint} & 130 \\
{Mathematical Programming} & \texttt{gurobipy, cvxpy, pyscipopt, statsmodel} & 76 \\
{Medicine} & \texttt{medcalc-bench} & 916 \\
{Security \& Safety} & \texttt{cryptography, gmpy2, pycryptodome} & 516 \\
{Programming} & - & 585 \\
\bottomrule
\end{tabular}
\end{table}

We begin by manually collecting domain-specific datasets consisting of questions and ground truth answers. Each question in the seed dataset is ensured to be solvable using code. If available, we also record the code that leads to the ground truth. The purpose of the seed dataset is not to be a large-scale dataset to use directly for training, but as a means to bootstrap the synthetic data generation process by seeding the generative process of the LLM.

To ensure broad coverage across diverse dataset types, we first collected 8,729 data points spanning 12 different domains to construct \loongbench. Detailed statistics of the \loongbench are provided in \Cref{tab:seed_dataset}. 
In particular, every data point in the seed set contains:
    \begin{itemize}
        \item a \emph{natural language question},
        \item a \emph{verified final answer}, and
        \item an \emph{accompanying rationale} in the form of executable python code.
        \item corresponding \emph{metadata}, including license, source, domain, required dependencies, name, contributor, creation date, difficulty level, and relevant tags.
    \end{itemize}

We start by introducing, in detail, how we collect the seed datapoints for each domain. 

\paragraph{Advanced Math}
To construct a high-quality seed dataset for advanced mathematical problem-solving, we begin by selecting problems from the training split of the MATH~\citep{hendrycksmath2021} dataset, focusing on those labeled with difficulty levels 4 or 5. For each selected problem, we prompt the \texttt{o3-mini} model~\citep{openai2025o3mini} to generate corresponding \texttt{SymPy} code intended to solve it. We then filter out any outputs that fail to produce executable code or yield incorrect solutions. The remaining \texttt{SymPy} programs are executed to produce numerical or symbolic results, which are then compared against the ground truth answers using MathVerifier\footnote{\url{https://github.com/camel-ai/camel/blob/master/camel/verifiers/math_verifier.py}}--a tool designed to parse and evaluate LaTeX-formatted mathematical expressions. Only those samples with verified correct solutions are retained and added to the seed dataset. We ended up collecting 1,611 problems.

\paragraph{Advanced Physics}
We select the physics problems from Scibench~~\citep{wang2024scibench} and Olympiadbench~\citep{he2024olympiadbench}. We asked \texttt{o3-mini}~\citep{openai2025o3mini} to generate \texttt{sympy} code which solves the problem. We execute the code and compare the result with the ground truth answer. 
The answer has two parts: numerical value and unit. Unit conversion and dynamic tolerance are used in the verification process.
unit conversion: If the response unit doesn't match with the ground truth unit, perform unit conversion to match them (e.g. $1$km v.s. $1000$m, $200,000$ vs $2*10^5$). We then compare the converted numerical value and unit to the ground truth ones.
dynamic tolerance: Adjust relative tolerance when comparing the numerical results if needed (the default tolerance is $0.01$, but sometimes the ground truth answer is given in lower tolerance, e.g. when ground truth is $1.3e+02$, we should allow the answers to sit within a difference of $0.05e+02$)
For each problem, we perform a maximum of three attempts to generate the correct rationale. The correct ones are included in the seed dataset.

\paragraph{Chemistry}
We extracted examples from the ChemistryQA dataset available on HuggingFace~\footnote{\url{https://huggingface.co/datasets/avaliev/ChemistryQA}}, which contains a wide range of chemistry reasoning question-answer pairs. We selected examples from this dataset that present conceptually clear and computationally relevant problems. When the original examples provided concise and understandable questions with well-formatted answers, we retained them directly. For examples where the format was unclear or the answer presentation lacked structure, we reformulated them using the \texttt{o3-mini}~\citep{openai2025o3mini} model. This model allowed us to generate well-structured questions that emphasize general chemistry reasoning, such as stoichiometry, thermodynamics, or molecular structure, while avoiding references to implementation-specific details. For instance, an original question about molar mass calculation was rewritten as: \emph{``What is the molar mass of sulfuric acid ($H_2SO_4$)?''} This process resulted in 3076 seed data points, each aligned with core chemistry concepts and designed to support both human and model-based reasoning.

\paragraph{Computational Biology}
We collected datasets from the Biochemistry, Genetics, Molecular Biology, and Microbiology domains under the Biology category on the General Reasoning website\footnote{\url{https://gr.inc}}. These datasets were then preprocessed using a custom script to merge and transform the raw data into a unified question-answering format.

We employed \texttt{GPT-4o}~\citep{openai2024gpt4o} to generate code-based solutions for each question. Subsequently, we filtered out instances where the generated code could not be executed or failed to address the problem correctly. To ensure the quality and validity of the resulting dataset, we further utilized \texttt{GPT-4o-mini}~\citep{openai2024gpt4o} to evaluate whether:
(1) the generated code genuinely solved the problem rather than merely paraphrasing it;
(2) the execution result of the code was semantically consistent with the ground truth answer; 
(3) the code included necessary computational steps instead of directly hardcoding the output.
To more accurately evaluate whether the results of code execution align with the ground truth, we employed rule-based regular expression matching for both. Finally, we retained the question-answer data that showed consistent matches.
Only samples that passed all these criteria were included in the final seed dataset used for our experiments. We ended up collecting 51 samples. 

\paragraph{Finance}
We focus on extracting examples from a variety of QuantLib~\citep{quantlibpython2020} resources, including the official documentation\footnote{\url{quantlib-python-docs.readthedocs.io}}, online tutorials\footnote{\url{http://gouthamanbalaraman.com/blog/quantlib-python-tutorials-with-examples.html}}, and the FinAI Financial Reasoning Dataset on Hugging Face\footnote{\url{https://huggingface.co/TheFinAI}}. From these, we identified instances that contain well-defined financial modeling problems and accompanying code examples. When the original examples included clearly presented questions and answers, we retained them directly. For those that lacked clarity or structure, we used the \texttt{o3-mini} model~\citep{openai2025o3mini} to reformulate them into precise question formats that emphasize financial reasoning and conceptual understanding, while avoiding low-level implementation details. For instance, an example involving bond pricing using QuantLib was rewritten into the question: \emph{``What is the clean price of a fixed-rate bond with a face value of \$1000, a 5\% annual coupon, and 10 years to maturity if the yield is 4.5\%?''} This process yielded a curated set of seed data points suitable for financial reasoning tasks, each paired with a corresponding Python-based solution.

\paragraph{Board Game}
We adopt \textit{Blackjack} as the canonical imperfect-information board game and generate interaction data through a hybrid pipeline that combines reproducible simulation with quality-controlled traces derived from \textit{Agent Pro}~\citep{zhang2024agent}.
All fresh episodes are played in the {\small\textsc{RLCard}} environment~\citep{zha2020rlcard} against two fixed baseline opponents--Deep Q-Network (DQN)~\citep{mnih2015human} and Deep Monte-Carlo (DMC)~\citep{zha2021douzero}--so that the strategic background remains constant.
We execute games until we collect $50$ distinct \emph{losing} rounds; these serve as the seed corpus for policy-level reflection.
Following the protocol of \citet{zhang2024agent}, every state-action pair is augmented with (i) a natural-language rationale extracted from an expert Blackjack playbook and (ii) a behaviour guideline produced by the reflection mechanism.

To suppress hallucination and retain only strategically sound trajectories, we run perfect-information Monte-Carlo roll-outs~\citep{Arjonilla_2024} from each decision point to estimate the true probability of winning for every legal action.
A trajectory is kept only if the chosen action belongs to the top-$k$ actions ranked by this oracle probability (with $k{=}1$ in all experiments).
This filtering step aligns the sampled behaviour with optimal-play statistics and yields a high-fidelity corpus for subsequent safety and alignment analysis. We ended up generating 926 questions as described.

\paragraph{Graph \& Discrete Math}
We focus on extracting examples from the documentation of \texttt{networkx}\footnote{\url{https://networkx.org/}}~\citep{networkx} (abbreviated as \texttt{nx}), a widely used library for network science. We began by inspecting all 896 functions under \texttt{nx.algorithms}. From these, we identified 370 functions that included usage examples and extracted the corresponding code snippets as rationales. To ensure reproducibility, we filtered out examples involving multiple valid outputs or random sampling. The remaining examples were rewritten using \texttt{GPT-4o-mini}~\citep{openai2024gpt4o} to formulate questions that reflect the exact outcome of the code in general graph-theoretic terms, avoiding references to implementation details. For instance, the code snippet \texttt{nx.is\_k\_regular(nx.Graph([(1, 2), (2, 3), (3, 4), (4, 1)]), k=3)} was transformed into the question: \emph{``For a graph defined by the edges connecting nodes as follows: (1, 2), (2, 3), (3, 4), and (4, 1), is the graph 3-regular?''} This process yielded 178 seed data points, each corresponding to a distinct algorithm implementation in the \texttt{networkx} library. All data points were then manually checked to ensure that the code executes correctly and produces the intended result.

\paragraph{Logic} 

We currently have two types of constraint satisfaction problems (CSPs): Einstein's Puzzle\footnote{\url{https://en.wikipedia.org/wiki/Zebra_Puzzle}} and Analytical Reasoning questions from the Law School Admission Test (AR-LSAT)\footnote{\url{https://www.lsac.org/lsat}}.

Einstein's Puzzle is a deductive reasoning task where a set of positions must each be assigned a unique combination of items across several categories, guided by a set of interrelated clues (constraints). Solutions can be represented as dictionaries, with categories as keys and ordered lists of items as values. To generate these puzzles, we define a pool of 10 possible categories (e.g., \texttt{Job}, \texttt{Beverage}, \texttt{Food}) and 25 items per category (e.g., \texttt{Accountant} and \texttt{Doctor} 
 belong to \texttt{Job}). 
For each instance, we randomly select a subset of categories and items, enumerate all possible permutations, and iteratively add constraints until only one valid solution remains. The final puzzle is constructed using a natural language template that integrates the scenario description, sampled categories, items, and constraints.

AR-LSAT questions assess logical reasoning within a structured system of relationships by requiring valid conclusions based on a set of rules and conditions. For example, a problem might involve scheduling presentations for \texttt{Mary}, \texttt{John}, and \texttt{Alice} on \texttt{Monday}, \texttt{Tuesday}, and \texttt{Wednesday}, with constraints such as \texttt{John} presenting after \texttt{Mary}. The task is to determine which conclusions follow logically. We use questions collected from \cite{zhong2021ar} as data points.

Since both Einstein's Puzzle and AR-LSAT are CSPs, we employ \texttt{GPT-4.1-mini}~\citep{openai2025gpt41} to generate Python code using the \texttt{python-constraint} library\footnote{\url{https://pypi.org/project/python-constraint/}}, ensuring each problem is solvable and has a unique solution.

\paragraph{Mathematical Programming}
The mathematical programming domains focus on solving optimization problems for an objective function that is subject to constraints. We mainly obtain questions directly from the existing tutorials from Gurobipy\footnote{\url{https://github.com/Gurobi/modeling-examples}}~\citep{gurobi}, PySCIPOpt\footnote{\url{https://github.com/scipopt/PySCIPOpt}}~\citep{MaherMiltenbergerPedrosoRehfeldtSchwarzSerrano2016}, CVXPY\footnote{\url{https://www.cvxpy.org/examples/}}~\citep{cvxpy}, and Statsmodels\footnote{\url{https://www.statsmodels.org/stable/examples/index.html}}~\citep{seabold2010statsmodels}. 
For all these datasets, we began by reviewing all official example notebooks provided in their corresponding documentation and identified a subset that met our criteria for practical relevance and technical clarity. Specifically, we selected notebooks that included detailed real-world problem descriptions, utilized data from identifiable sources, and were compatible with the latest Jupyter environment. From these, we curated a collection of problems where the accompanying data were either regenerated using \texttt{GPT-4o}~\citep{openai2024gpt4o}--guided by the original sources--or partially retained to preserve solvability. Markdown formatting issues were corrected, and scripts were streamlined by removing non-essential outputs such as extraneous \texttt{print} statements and visualizations. Logical inconsistencies and errors within the solution code were minimally edited, and final answers were marked using the \texttt{\textbackslash boxed{}} notation for clarity.

\paragraph{Medicine}
We constructed our dataset from the MedCalc-Bench benchmark\footnote{\url{https://ncbi/MedCalc-Bench-v1.0}} through the following procedure. We start by merging the ``patient note'' and ``question'' columns into a unified question field, and designating the ``Ground Truth Answer'' column as the ``final answer''.
To ensure diversity across categories, we sampled up to 30 entries for each unique ``Calculator Name.'' For calculators with fewer than 30 available entries, we included all available data. This yielded a total of 1,192 examples.

We then improved the official tool code~\citep{khandekar2024medcalcbenchevaluatinglargelanguage} by adding docstrings and performing additional code optimizations. The entire codebase was then consolidated into a Python package and also stored in a single text file. We prompted \texttt{GPT-4.1}~\citep{openai2025gpt41} to generate code (as rationale) capable of solving each question using functions from the medcalc-bench package. 
We filtered the data by discarding any question-answer pairs for which the generated rationale code failed to execute successfully. This filtering step resulted in 1,117 valid examples. We then compared the outputs of the executed rationales with the corresponding ``final answer'' fields using a rule-based regular expression matching system. Through this process, we identified 916 question-answer pairs with matched outputs. 

\paragraph{Security \& Safety}
We focused on extracting representative examples from existing cryptographic Capture-The-Flag (CTF) problems and their associated explanations. Specifically, we manually curated problems from the CTF-Wiki\footnote{\url{https://ctf-wiki.org/crypto/introduction/}}, which encompasses 3 major categories and 44 subcategories of cryptographic challenges. For each subcategory, we ensured coverage by including at least 3-5 representative problems.
To diversify and scale the dataset, we systematically replaced the plaintext and ciphertext in the original problems to generate multiple variants, ensuring each subcategory included at least 5 distinct instances. Where applicable, we used \texttt{GPT-4o}~\citep{openai2024gpt4o} to rewrite the textual components of the problems, making them more natural while preserving the cryptographic core. 

For each problem, we also collected the corresponding solution logic and reference implementation. With the assistance of \texttt{GPT-4o}~\citep{openai2024gpt4o}, we manually refined these into coherent rationales that explain the step-by-step solution process. All solution scripts were executed in a Python interpreter, and their outputs were compared against the official flags or expected results. Any examples with mismatched outputs were discarded.
The final dataset consists of problem statements, rationale code, and verified outputs for each data point. All entries were manually reviewed to ensure correctness, clarity, and reproducibility.

\paragraph{Programming}

The dataset taken from the LeetCode dataset~\citep{xia2025leetcodedatasettemporaldatasetrobust} and is constructed through a two-stage rollout procedure aimed at building a code critic model--an LLM designed to identify and fix errors in solutions to competitive programming problems. Each data point consists of a programming problem, an initial candidate solution, an automatic correctness judgment, and a potential correction. In the first round rollout, an initial candidate solution is generated using \texttt{Claude-3.5-Sonnet}~\citep{anthropic2024claude35sonnet} for each programming task. These candidate solutions then go through a second round evaluation where \texttt{DeepSeek-R1}~\citep{guo2025deepseekr1}. This model acts as an automated code critic: it first determines whether the initial solution is correct, outputting a binary true/false label. If the solution is found to be incorrect, the model attempts to revise it by producing a corrected version that better aligns with the problem specification.

\subsection{\loongenv: Modular Synthetic Generation with Verifiable Supervision}

\loongenv{} is a modular, black-box synthetic data generator seeded with a high-quality dataset, such as \loongbench. Given this seed, it can generate an unbounded number of question-answer pairs that expand the training distribution in a controllable and verifiable manner. The generator is abstracted from downstream modules, enabling the flexible integration of various prompting strategies and agent behaviors.

We support both simple prompting methods and complex multi-agent generation workflows. This section details the generation process and experimental configurations.

\paragraph{Question Synthesis}
We explore three strategies for generating synthetic questions from seed examples:

\begin{itemize}
    \item \textbf{Few-shot prompting}~\citep{brown2020language}: We provide a few seed QA pairs as demonstrations and prompt the model to generate new problems in a similar style. This serves as the simplest generation baseline.
    \item \textbf{Self-Instruct}~\citep{wang-etal-2023-self-instruct}: An instruction-tuned model is recursively prompted to generate increasingly diverse and structured prompts. 
    \item \textbf{Evol-Instruct}~\citep{xu2023wizardlm}: This approach evolves seed questions through mutation operations such as generalization, specification, and complexity scaling. 
\end{itemize}

\paragraph{Answer Synthesis}
For every generated question, we generate a corresponding answer using a coder agent, which generates the corresponding code and tries to execute it to obtain the results. While we do not assume the correctness of these synthetic answers, code execution enables us to produce grounded numerical outputs in many domains. Notably, we do not use the raw synthetic answers directly for training.


\paragraph{Verifiers}

To ensure high-quality generated data, we incorporate a verification mechanism that filters out incorrect synthetic solutions generated by our pipeline. To do this effectively, we validate synthetic answers using two independent approaches:

Deriving one solution directly through the Generator's code execution.
Independently generating another solution via natural-language Chain-of-Thought (CoT) reasoning.
If these independent solutions agree, it's highly likely that the answer is correct. Although rare, there's still a possibility of false positives (both approaches incorrectly agreeing). However, given the fundamentally different methods involved, we believe this will not occur often enough to be detrimental to model training.

Each environment also includes a verifier that semantically compares the LLM response with the synthetic answer, ensuring they are effectively equivalent. This verification step is crucial for accurately filtering semantic equivalences, significantly reducing false negatives (cases where semantically correct answers would otherwise be wrongly rejected).

Note that while our current setup relies on the \textit{LLM-as-a-judge} paradigm, where large language models are used to assess the correctness of solutions, we ultimately aim to develop domain-specific verifiers (e.g., for mathematics or programming), which are both more efficient and more reliable.

\paragraph{Future direction} In future work, we plan to use this verification framework to support reinforcement learning (RL). Specifically, the CoT-generating agent, the model we ultimately aim to train, can receive positive rewards only when its final answer is semantically verified to match the trusted synthetic answer. This setup enables a Reinforcement Learning from Verified Rewards (RLVR) paradigm, where the agent learns exclusively from high-confidence, semantically aligned supervision.

\section{Experiments}

We evaluate the capabilities of state-of-the-art language models on \loongbench~and synthetic datasets produced by \loongenv. Specifically, we aim to answer the following questions:
\begin{itemize}
    \item How well models perform across diverse reasoning domains on \loongbench?
    \item How reliably can they generate executable and semantically valid solutions using \loongenv?
    \item How do different prompting strategies affect the diversity and difficulty of the generated tasks?
\end{itemize}

\subsection{Setup}
\label{sec:expset}

\paragraph{Dataset and Models.} We evaluate the reasoning and problem-solving capabilities of a range of state-of-the-art language models across 12 domains provided as \loongbench: \textbf{Advanced Math}, \textbf{Advanced Physics}, \textbf{Chemistry}, \textbf{Computational Biology}, \textbf{Finance}, \textbf{Game}, \textbf{Graph \& Discrete Math}, \textbf{Logic}, \textbf{Mathematical Programming}, \textbf{Medicine}, \textbf{Security \& Safety}, and \textbf{Programming}. 
Each domain consists of curated problems sourced from high-quality benchmarks, academic competitions, and domain-specific datasets, as described in \Cref{sec:seeddataset}. 
We include both open- and closed-source language models in our evaluation. Proprietary models such as \texttt{GPT4.1-mini}~\citep{openai2025gpt41}, \texttt{o3-mini}~\citep{openai2025o3mini}, \texttt{Grok-3}~\citep{xai2025grok3}, and \texttt{Claude-3.7-Sonnet}~\citep{anthropic2025claude37} provide strong baselines from leading commercial providers. In parallel, we incorporate high-performing open-source models like \texttt{DeepSeek-r1}~\citep{guo2025deepseekr1} and \texttt{Qwen3-8B}~\citep{qwen3}, both of which demonstrate competitive performance on reasoning-intensive tasks.

\paragraph{Implementation.} All models are evaluated using their publicly available APIs or checkpoints, with consistent temperature, top-$k$, and top-$p$ settings where applicable. For fairness, we disable function/tool calling unless stated otherwise, and restrict outputs to a single response per prompt without retries. We set the max-token for all experiments to 4096\footnote{Some questions in \textbf{mathematical programming} thus go out of this window and the result would be incomplete.}. We use a single NVIDIA H100 80GB GPU for inference with the open-sourced model. Our development is based on the CAMEL framework~\citep{li2023camel}, and the codebase supporting this work is publicly available at \url{https://github.com/camel-ai/loong}.

\paragraph{Evaluation.}
Accuracy is measured as the percentage of correctly answered problems per domain. We employ LLM-as-judge~\citep{gu2025surveyllmasajudge}, using a \texttt{GPT4.1-mini}~\citep{openai2025gpt41}, to assess correctness due to the complex nature of answers, which may vary in format across different domains. The judge accounts for symbolic equivalence where applicable, ensuring that mathematically correct but differently expressed answers are not penalized.

\begin{table}[!t]
\small
\centering
\caption{Benchmarking accuracy across domains and model categories. The best model is highlighted in \textbf{bold}, and the second best is \underline{underlined}.}
\label{tab:domains_by_model}
\begin{tabular}{{l}|*{6}{>{\centering\arraybackslash}p{1.7cm}}}
\toprule
\textbf{Domain}
  & {\scriptsize \texttt{GPT4.1-mini}}
  & {\scriptsize \texttt{o3-mini}}
  & {\scriptsize \texttt{Grok-3}}
  & {\scriptsize \texttt{Claude-3.7}}
  & {\scriptsize \texttt{DeepSeek-r1}}
  & {\scriptsize \texttt{Qwen3-8B}} \\
\midrule
Advanced Maths           & 91.4 & \textbf{97.4} & 92.3 & 79.3 & \underline{96.7} & 79.2 \\
Advance Physics          & 71.8 & \underline{75.3} & 69.0 & 63.9 & \textbf{77.4} & 59.2 \\
Chemistry          & 75.2 & 79.5 & 71.2 & \textbf{80.7} & 74.7 & \underline{79.7} \\
Computational Biology     & \underline{90.2} & 88.2 & \textbf{96.1} & \underline{90.2} & 88.2 & 86.2 \\
Finance           & 23.8 & \textbf{24.3} & 19.1 & 22.0 & \textbf{24.3} & 12.8 \\
Game           & 92.0 & \underline{96.0} & 93.0 & 95.1 & \textbf{97.3} & 43.2 \\
Graph          & 80.9 & \underline{82.0} & 80.1 & 73.6 & \textbf{83.7} & 62.9 \\
Logic          & \textbf{65.4} & 61.6 & 55.4 & 46.9 & \underline{62.3} & 39.2  \\
Math. Programming    & \underline{11.8} & 9.2  & 6.4  & \textbf{13.2} & 10.5 & 10.0 \\
Medicine           & \textbf{59.6} & 46.3 & 50.7 & 54.1 & 52.6 & 28.4 \\
Security       & \underline{25.6} & 11.2 & 22.3 & 4.7  & \textbf{28.7} & 7.9 \\
Programming          & 98.6 & \textbf{100.0} & 91.5 & 97.4 & \underline{98.8} & 81.7 \\
\bottomrule
\end{tabular}
\end{table}

\subsection{Benchmarking \loongbench}

We report the benchmarking results in \Cref{tab:domains_by_model}.
The results reveal key trends regarding domain difficulty, model specialization, and the performance gap between open- and closed-source systems. We highlight three central findings below.

\paragraph{A well-calibrated spectrum of difficulty.}
The twelve domains in our benchmark exhibit a wide range of difficulty levels. For example, the \emph{Mathematical Programming} domain yields accuracies as low as 10\%, indicating substantial unresolved complexity, whereas the \emph{Programming} domain is nearly saturated, with models like \texttt{o3-mini} achieving 100\% accuracy. Other domains, such as \emph{Logic}, \emph{Graph \& Discrete Math}, and \emph{Chemistry}, distribute evenly across the middle range. This balance ensures the benchmark is broadly discriminative: it avoids ceiling effects for strong models while remaining accessible for lower-capacity systems, making it a robust testbed for both evaluation and ablation studies.

\paragraph{Reasoning-optimized models consistently outperform.}
We observe that models explicitly tuned or pretrained for reasoning, particularly \texttt{o3-mini} and \texttt{DeepSeek-r1}, achieve top scores across the majority of domains. Notably, \texttt{DeepSeek-r1} consistently reached top-2 in 8 out of 12 datasets, whereas \texttt{o3-mini} also reached 6 out of 12. These results suggest that the benchmark requires more than surface-level pattern matching or factual recall: it emphasizes structured, multi-step reasoning, which is better captured by models with strong CoT or planning capabilities.

\paragraph{Open-source models lag in reasoning-heavy domains.}
Although open models like \texttt{DeepSeek-r1} and \texttt{Qwen3-8B} perform competitively in certain areas, a clear performance gap remains in the most reasoning-intensive domains. For instance, in the \emph{Game} and \emph{Logic} domains, \texttt{Qwen3-8B} trails \texttt{o3-mini} by 50 and 22 percentage points, respectively. These discrepancies highlight two key challenges: (i) current open-source systems underperform on strategy-based and logic-heavy tasks, and (ii) this benchmark suite is well-positioned to reveal such fine-grained capability gaps, offering concrete targets for future model development and alignment in the open-source community.

\subsection{Synthesizing Data with \loongenv}

We use \loongenv~to generate synthetic data under three instruction paradigms: \textbf{Few-shot prompting}~\citep{brown2020language}, \textbf{Self-instruct}~\citep{wang-etal-2023-self-instruct}, and \textbf{Evol-instruct}~\citep{xu2023wizardlm}.
%
For each setting, we initialize the generator with seed examples from \loongbench~and use a multi-agent workflow to synthesize new data:
\begin{enumerate}
    \item A \textbf{question synthesis agent} generates a new natural language question based on the seed data.
    \item A separate \textbf{code generation agent} is then prompted to produce an executable program that answers the generated question.
\end{enumerate}

We then evaluate the quality of the generated samples in two stages:

\begin{itemize}
    \item \textbf{Executability Check:} We run each generated code snippet in a sandboxed Python environment and measure the fraction that executes without error. This provides a proxy for functional correctness and yields the \emph{execution success rate}.
    
    \item \textbf{Verification via Judge Agent:} For each generated question-code pair, we prompt a \textbf{judge agent} to assess two criteria: (1) whether the question is well-formed and meaningful, and (2) whether the generated code correctly solves the posed problem. The fraction of samples that fail either criterion determines the \emph{rejection rate}.
\end{itemize}

This two-stage evaluation process provides a systematic measure of both functional correctness and semantic fidelity of the generated question-code pairs. 

\paragraph{Implementation.} We use \texttt{GPT-4.1-mini}~\citep{openai2025gpt41} as both the question synthesis agent and the code generation agent across all experiments. For each domain and each generation strategies, we randomly selected from the seed dataset and generated 100 synthetic questions. For verification, we employ \texttt{DeepSeek-R1}~\citep{guo2025deepseekr1} as the judge agent. All other settings remain consistent with the experimental setup described in the previous section.

\subsubsection{Execution and Verification Outcomes}

\begin{figure}[t]
\centering
\begin{tikzpicture}
\begin{axis}[
    ybar stacked,
    bar width=0.3cm,
    width=10cm,
    height=6cm,
    enlarge x limits=0.2,
    ylabel={Percentage (\%)},
    symbolic x coords={
        FewShot-L, Self Instruct-L, Evol Instruct-L,
        ,
        FewShot-P, Self Instruct-P, Evol Instruct-P},
    xtick=data,
    xticklabels={
        FewShot, Self-Ins, Evol-Ins, ,
        FewShot, Self-Ins, Evol-Ins
    },
    xticklabel style={
    font=\tiny, draw=none},
    ymin=0, ymax=100,
    grid=both,
    ymajorgrids=true,
    minor y tick num=1,
    legend style={
        at={(0.5,-0.18)},
        anchor=north,
        legend columns=-1,
        font=\scriptsize
    }
]

\addplot+[fill=blue!30] coordinates {
    (FewShot-L,92.6) (Self Instruct-L,53.8) (Evol Instruct-L,41.6)
    (,0)
    (FewShot-P,93.9) (Self Instruct-P,82.0) (Evol Instruct-P,56.2)};
\addplot+[fill=orange!30] coordinates {
    (FewShot-L,4.6) (Self Instruct-L,44.8) (Evol Instruct-L,3.5)
    (,0)
    (FewShot-P,4.7) (Self Instruct-P,3.3) (Evol Instruct-P,29.8)};
\addplot+[fill=red!30] coordinates {
    (FewShot-L,2.8) (Self Instruct-L,1.3) (Evol Instruct-L,55.0)
    (,0)
    (FewShot-P,1.4) (Self Instruct-P,14.8) (Evol Instruct-P,14.0)};

\legend{Pass, Judge-Rejected, Not executable}
\end{axis}

\node at (2.05,5.0) {\textbf{Logic}};
\node at (6.1,5.0) {\textbf{Physics}};
\end{tikzpicture}
\caption{Execution outcome breakdown across synthetic data generation strategies for Logic and Physics domains.}
\label{fig:logic-physics-stacked}
\end{figure}
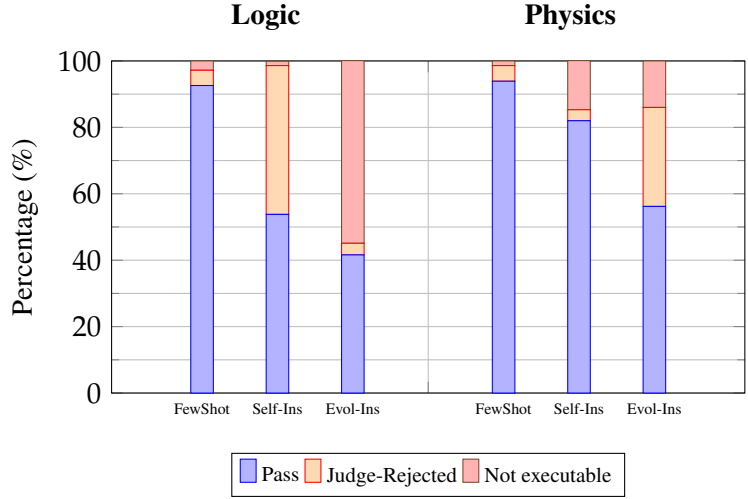

\Cref{fig:logic-physics-stacked} summarizes the execution outcomes of synthetic data generated by \loongenv~under three prompting strategies-FewShot, Self-Instruct, and Evol-Instruct-across two domains: \emph{Logic} and \emph{Physics}. We categorize outcomes into three classes: \emph{Pass} (code executed and the judge approved the answer), \emph{Judge-Rejected} (code executed but result disagreed with the judge's answer), and \emph{Not Executable} (code failed to run).

We observe that in the \emph{Logic} domain, FewShot prompting yields a high pass rate (92.6\%) with very few failures, while Self-Instruct exhibits a much higher rejection rate (44.8\%) and Evol-Instruct produces a large portion of non-executable code (55.0\%). In contrast, for the \emph{Physics} domain, both FewShot and Self-Instruct maintain high pass rates (93.9\% and 82.0\% respectively), but Evol-Instruct again suffers from significantly reduced executability and semantic agreement, with 29.8\% judged as incorrect and 14.0\% failing to execute.

These results highlight a trade-off between prompting complexity and generation reliability: while FewShot prompting offers the most stable pipeline with the highest overall pass rates, Evol-Instruct, despite its higher rejection and execution failure rates, remains highly valuable from a training perspective. Its ability to synthesize more diverse and challenging reasoning tasks makes it especially well-suited for building robust models. As we demonstrate in the later sections, Evol-Instruct more effectively captures edge cases and reasoning depth that are essential for meaningful generalization.

\subsubsection{Diversity}

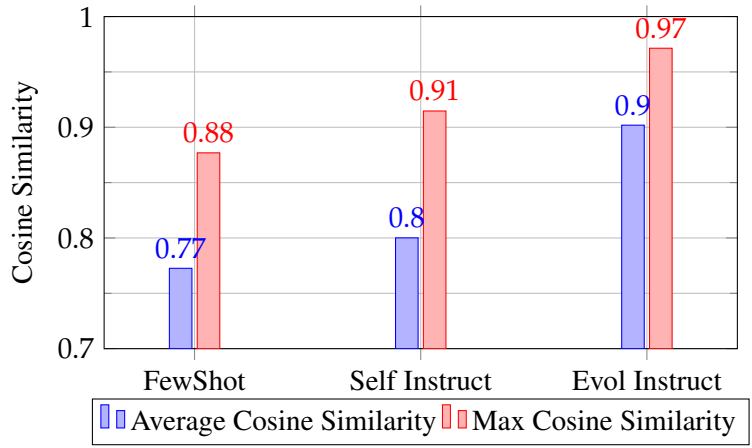
\begin{figure}[t]
\centering
\begin{tikzpicture}
\begin{axis}[
    ybar,
    bar width=0.3cm,
    width=10cm,
    height=6cm,
    enlarge x limits=0.2,
    ylabel={Cosine Similarity},
    symbolic x coords={FewShot, Self Instruct, Evol Instruct},
    xtick=data,
    ymin=0.7, ymax=1.0,
    legend style={at={(0.5,-0.15)}, anchor=north, legend columns=-1},
    nodes near coords,
    nodes near coords align={vertical},
    grid=both,
    ymajorgrids=true,
    minor y tick num=1
]

\addplot+[style={blue, fill=blue!30}] 
    coordinates {(FewShot, 0.7726) (Self Instruct, 0.8001) (Evol Instruct, 0.9019)};
\addplot+[style={red, fill=red!30}] 
    coordinates {(FewShot, 0.8769) (Self Instruct, 0.9147) (Evol Instruct, 0.9714)};

\legend{Average Cosine Similarity, Max Cosine Similarity}
\end{axis}
\end{tikzpicture}
\caption{Seed-generated pairwise cosine similarity on the Advanced Physics domain across synthetic data generation strategies.}
\label{fig:cosine-similarity-physics}
\end{figure}

\begin{figure*}[t]
\centering
\begin{minipage}{0.32\textwidth}
    \centering
    \includegraphics[width=\linewidth]{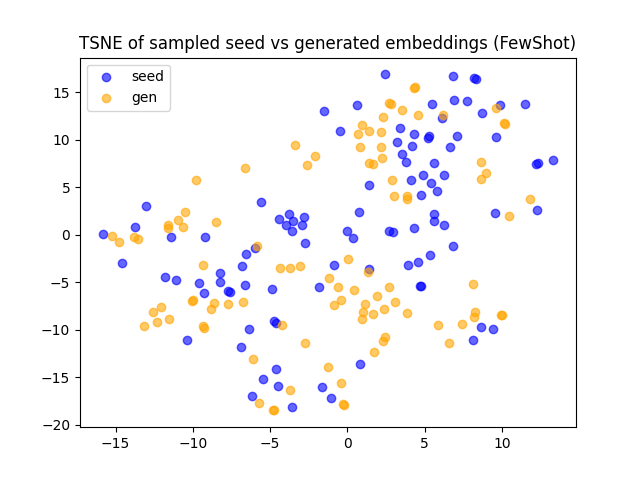}
    \caption*{(a) Few-shot}
\end{minipage}
\hfill
\begin{minipage}{0.32\textwidth}
    \centering
    \includegraphics[width=\linewidth]{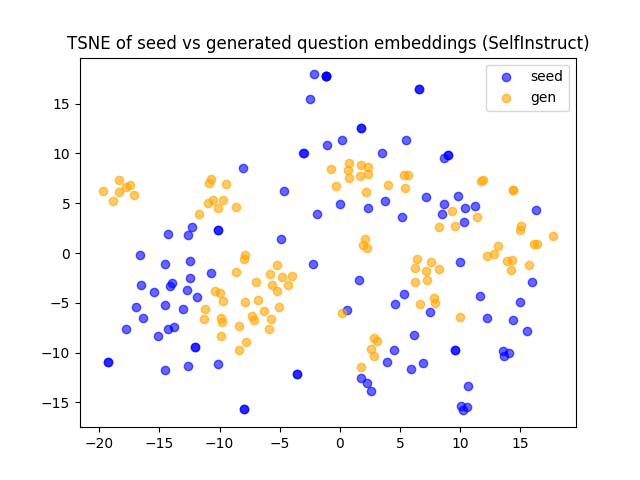}
    \caption*{(b) Self-Instruct}
\end{minipage}
\hfill
\begin{minipage}{0.32\textwidth}
    \centering
    \includegraphics[width=\linewidth]{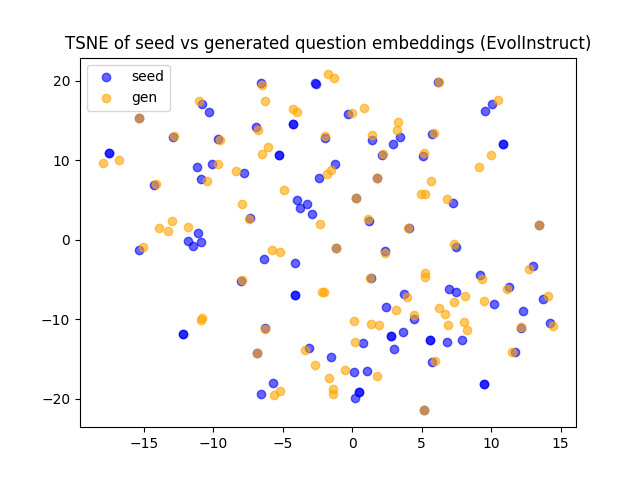}
    \caption*{(c) Evol-Instruct}
\end{minipage}
\caption{t-SNE projection of embedding space for seed vs.\ generated problems on the Advanced Physics across different generation strategies. Generated examples (orange) and seed samples (blue) cluster with varying degrees of overlap, indicating distributional proximity and diversity.}
\label{fig:tsne-comparison}
\end{figure*}

\begin{figure*}[t]
\centering
\begin{minipage}{0.32\textwidth}
    \centering
    \includegraphics[width=\linewidth]{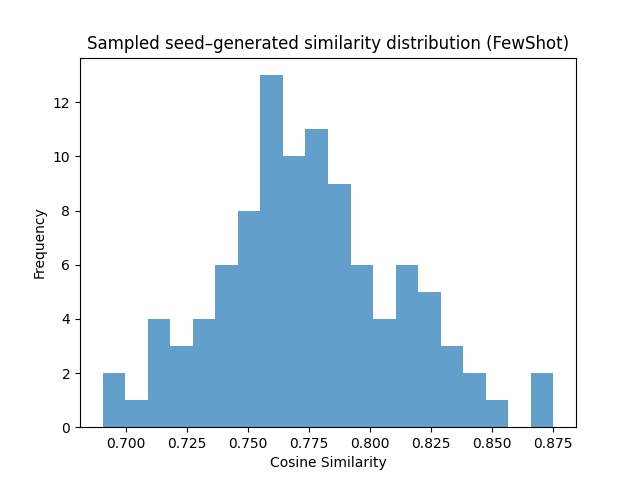}
    \caption*{(a) Few-shot}
\end{minipage}
\hfill
\begin{minipage}{0.32\textwidth}
    \centering
    \includegraphics[width=\linewidth]{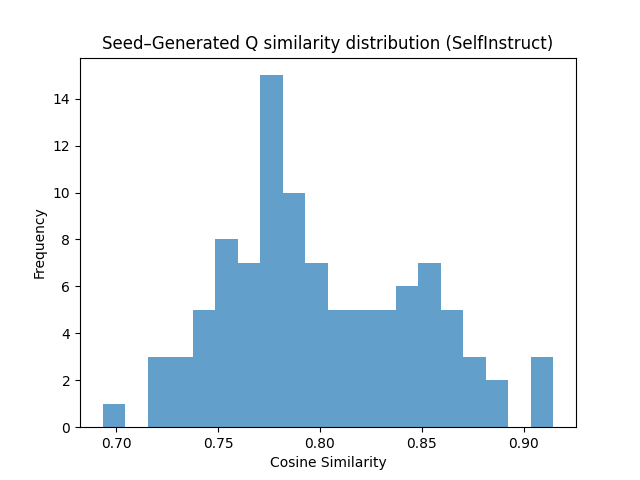}
    \caption*{(b) Self-Instruct}
\end{minipage}
\hfill
\begin{minipage}{0.32\textwidth}
    \centering
    \includegraphics[width=\linewidth]{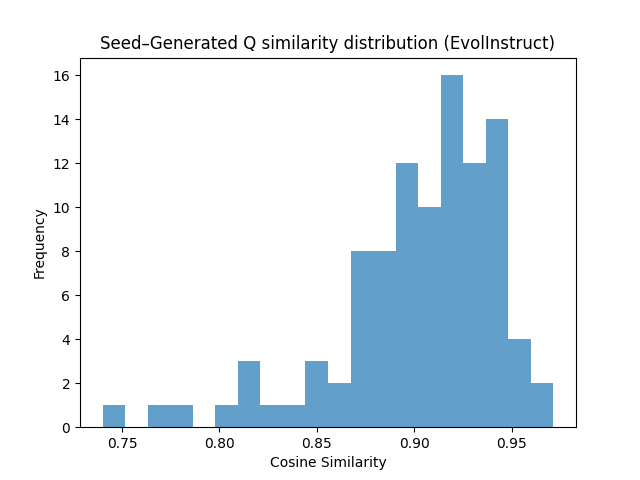}
    \caption*{(c) Evol-Instruct}
\end{minipage}
\caption{Cosine similarity distribution between seed and generated questions on the Advanced Physics domain for different generation strategies.}
\label{fig:cosine-similarity-hist}
\end{figure*}

We assess the semantic diversity of generated questions in the Advanced Physics domain by comparing their embeddings to those of the seed questions using cosine similarity and t-SNE visualization. \Cref{fig:cosine-similarity-physics} reports the average and maximum cosine similarity between 100 seed-generated question pairs under each strategy, while \Cref{fig:tsne-comparison,fig:cosine-similarity-hist} provide a qualitative breakdown of embedding proximity using t-SNE and distributional skew.

Overall, we observe that \textbf{Few-Shot} prompting tends to produce questions that are more lexically distinct from the seeds, as reflected by its lower average similarity (0.77) and more dispersed t-SNE distribution. However, these questions often remain close to the original in structure and complexity, representing surface-level variation.

In contrast, \textbf{Self-Instruct} generates questions that are both lexically and semantically more diverse, often drifting farther from the seed set in the t-SNE space. This suggests a bias toward novelty, albeit sometimes at the expense of coherence or executability (see correctness analysis).

Interestingly, \textbf{Evol-Instruct} generates questions that appear much closer to the seed questions semantically, as suggested by the high average (0.90+) and maximum cosine similarities, along with tight clustering in the t-SNE plots. This pattern may indicate that Evol-Instruct tends to apply transformations, such as generalization, specification, or rephrasing, that preserve the underlying semantics while increasing complexity.

These observations suggest a potential trade-off between surface-level lexical variation and semantic alignment: while Few-Shot prompting yields more lexically distinct outputs, Evol-Instruct may generate structurally richer and semantically coherent examples. We hypothesize that such properties could be beneficial for capturing deeper reasoning patterns and edge cases, which are valuable for robust model training.

\subsubsection{Difficulty}

Finally, we proceed to measure the difficulty level of the generated datasets. Here, we again focus on \emph{Advanced Physics} domain and assess the difficulty of generated questions by measuring the accuracy of two models: \texttt {GPT4.1-mini} and \texttt{DeepSeek-r1}.

\Cref{tab:physics_accuracy} presents model accuracies on questions generated via each strategy in the Advanced Physics domain. We observe that both \texttt{GPT4.1-mini} and \texttt{DeepSeek-r1} perform best on Few-Shot generated data, achieving 92.0\% and 93.2\% accuracy respectively. Accuracy slightly drops on Self-Instruct data (83.0\% and 87.4\%), and significantly declines on Evol-Instruct questions (62.0\% and 70.3\%).

Interestingly, despite the Evol-Instruct examples being more semantically similar to the seed questions (\Cref{fig:cosine-similarity-physics}), their lower model accuracy suggests that they are substantially harder to solve. This supports our earlier hypothesis that Evol-Instruct tends to preserve core semantics while increasing reasoning complexity, possibly through abstract transformations or compound formulations.

\begin{table}[t]
\centering
\caption{Average accuracy (\%) on Advanced Physics domain across synthetic data generation strategies}
\label{tab:physics_accuracy}
\begin{tabular}{lccc|c}
\toprule
\textbf{Model} & \textbf{Few-shot} & \textbf{Self-Instruct} & \textbf{Evol-Instruct} & \textbf{Seed Dataset} \\
\midrule
\texttt{GPT4.1-mini}   & 92.0 $\uparrow$ & 83.0 $\uparrow$ & 62.0 $\downarrow$ & 71.8 \\
\texttt{DeepSeek-r1}   & 93.2 $\uparrow$ & 87.4 $\uparrow$ & 70.3 $\downarrow$ & 77.4 \\
\bottomrule
\end{tabular}
\end{table}



\section{Related Work}

\paragraph{Post-training for LLM}
Early efforts in aligning large language models with human preferences focused on fine-tuning via human feedback. InstructGPT demonstrated that reinforcement learning from human feedback (RLHF) can make smaller models more helpful, honest, and harmless than much larger base models \citep{leike2022training}, building on earlier work that optimized from pairwise preferences \citep{christiano2017deep} and fine-tuned models to match human judgments \citep{ziegler2019fine}.
Recent approaches pursue more stable or efficient alignment techniques. Direct Preference Optimization (DPO) reframes reward modeling as a supervised objective \citep{rafailov2023dpo}, while Preference Ranking Optimization (PRO) models task-aware ranks \citep{song2023pro} and RRHF bypasses RL entirely \citep{yuan2023rrhf}. Other efforts include training helpful assistants via RLHF \citep{leike2022training}, improving summarization through feedback \citep{stiennon2020learning}, and formalizing verifiable reward learning \citep{mroueh2025grpo}. Co-evolutionary training with unit testers \citep{wang2025coevollm}, single-example reward tuning \citep{wang2025oneshotrlvr}, autoregressive search \citep{shen2025satori}, cross-domain RLVR \citep{su2025expandingrlvr}, and prolonged fine-tuning \citep{liu2025prorl} further extend the design space.

\paragraph{Synthetic data generation}
In recent years, there has been a surge of interest in using language models to synthesize their own training data. Self-Instruct introduced a pipeline in which models generate and train on synthetic instructions to improve alignment \citep{wang2022selfinstruct}. WizardLM \citep{xu2023wizardlm} and WizardCoder \citep{luo2023wizardcoder} extend this with Evol-Instruct, automatically evolving instruction complexity for general and code tasks, respectively. The Flan Collection \citep{longpre2023flancollection} curates diverse instructional tasks and data augmentation strategies to boost zero-shot generalization, while LIMA \citep{zhou2023lima} shows that only a small amount of high-quality prompts can yield strong performance. Super-NaturalInstructions \citep{wang2022super} aggregates over 1600 NLP tasks to study generalization to unseen instructions.
Beyond instruction generation, more systematic data generation frameworks have emerged. DataGen provides a unified pipeline with modules for controllability, diversity, and factuality \citep{huang2025datagen}. A comprehensive survey reviews challenges and advances in LLM-based synthetic data generation for text and code \citep{nadas2025syntheticsurvey}.

\paragraph{Reinforcement Learning with Verifiable Reward (RLVR)}
RLVR combines reinforcement learning with automatic, verifiable reward signals, often programmatic correctness or agreement with auxiliary tools, to scale alignment and reasoning. GRPO formalizes reward dynamics \citep{mroueh2025grpo}, while one-shot RLVR shows that even a single example can substantially improve reasoning \citep{wang2025oneshotrlvr}. Satori proposes a Chain-of-Action-Thought framework \citep{shen2025satori}, and ProRL shows that prolonged optimization enables emergent capabilities \citep{liu2025prorl}.
RLVR has also been applied to instruction following \citep{su2025expandingrlvr}, co-evolutionary coding setups \citep{wang2025coevollm}, and multi-domain generalization. These efforts enable scalable, automatic reward supervision in open-ended environments \citep{liu2025prorl, su2025expandingrlvr, wang2025coevollm, shen2025satori, wang2025oneshotrlvr, mroueh2025grpo}.


\section{Conclusion}

We present \loong Loong, a modular framework for aligning LLMs via synthetic data generation and verifiable reward supervision. Our approach is driven by two key insights: effective alignment requires diverse, domain-specific data and scalable, annotation-free reward mechanisms.

Our contributions are fourfold: (1) \textbf{\loongbench}, a seed dataset of 8,729 examples across 12 reasoning-intensive domains with executable code and verified answers; (2) \textbf{\loongenv}, a flexible environment enabling diverse synthetic data generation strategies; (3) comprehensive benchmarking of open-source and proprietary models to assess domain generalization; and (4) in-depth analysis of generated data quality in terms of correctness, diversity, and complexity.

Together, these components form a cohesive framework for studying alignment at scale. Our results demonstrate that structured synthetic generation and verification can yield diverse and challenging synthetic datasets. A key future direction is leveraging \loongenv\  to support RLVR with synthetically generated questions. We also plan to extend \loongenv with tool-augmented generation and formal abstraction, and scale \loongbench to cover multilingual and multimodal tasks.

\section*{Acknowledgements}
We sincerely thank  Weijie Bai, Yuqin Xie and Yueming Lai for their valuable support in this project. We also thank Chen Li in SpiNNcloud Systems GmbH for his support on Sudoku problems. This work was carried out as a collaborative open source research initiative at
CAMEL-AI.org, supported by funding from Eigent.AI.

\newpage

\bibliographystyle{unsrtnat}
\bibliography{references} 
\clearpage
\appendix

\section{System Prompt for Benchmarking Experiments}
\label{appendix:prompts}

We use tailored prompts to standardize the generation and evaluation processes across all models tested in our benchmark. These prompts are described below.

\begin{tcolorbox}[colback=gray!5!white, colframe=black!75, title=Solver System Prompt, fonttitle=\bfseries]
\scriptsize
\texttt{You are an \{\{DOMAIN\}\} solver. It is imperative that you end with your answer wrapped in a \textbackslash boxed\{\} statement. It should only be the final answer, e.g., correct: \textbackslash boxed\{\textbackslash frac\{1\}\{4\}\}, incorrect: \textbackslash boxed\{x = \textbackslash frac\{1\}\{4\}\}.}
\end{tcolorbox}

\begin{tcolorbox}[colback=gray!5!white, colframe=black!75, title=LLM-as-Judge System Prompt, fonttitle=\bfseries]
\scriptsize
\texttt{You are an answer judge. Your task is the following: You will be provided with a GROUND TRUTH ANSWER, and with an entire answer from an LLM output. At the end of the LLM output, there should be a \textbackslash boxed\{\} statement containing an answer.}

\texttt{Your task is to compare this answer to the GROUND TRUTH ANSWER. You should only focus on the boxed statement at the end of the LLM response, disregard any reasoning steps before it.}

\texttt{You should end your response with your judgement wrapped in a \textbackslash boxed\{\} statement.}

\texttt{Wrap your evaluation result as \textbackslash boxed\{1\} if the provided LLM answer is mathematically equivalent (EVEN IF THE FORMAT IS DIFFERENT!!) to the GROUND TRUTH ANSWER. Wrap it as \textbackslash boxed\{0\} if the provided LLM answer is either not existent or not mathematically equivalent.}
\end{tcolorbox}

These prompts ensure consistency in both generation and evaluation phases and are critical to maintaining reproducibility and fairness in our benchmarking pipeline.


\end{document}